\documentclass[times,11pt,final]{elsarticle}

\usepackage{jcomp}
\usepackage{float}
\usepackage{amsfonts,amsmath,amsthm}
\usepackage[usenames,dvipsnames,svgnames,table,rgb]{xcolor}
\usepackage{graphicx,color} 
\usepackage{pdfpages}
\usepackage{graphicx}
\usepackage{subcaption} 
\usepackage{diagbox}
\usepackage{hyperref,float} 
\usepackage{tikz}
\usetikzlibrary{arrows.meta}
\usepackage{booktabs}
\usepackage{graphicx}

\definecolor{darkteal80}{RGB}{173,216,230}
\journal{Journal of Computational Physics}

\begin{document}

\begin{frontmatter}



\title{PO-CKAN:Physics Informed Deep Operator Kolmogorov Arnold Networks with Chunk Rational Structure}

\author[label1]{Junyi Wu}
\author[label1,label2,cor1]{Guang Lin} 

\cortext[cor1]{Corresponding author: Guang Lin, E-Mail: Guanglin@purdue.edu}

\affiliation[label1]{organization={Department of Mathematics, Purdue University},
            addressline={610 Purdue Mall}, 
            city={West Lafayette},
            postcode={47907}, 
            state={IN},
            country={USA}}
\affiliation[label2]{organization={School of Mechanical Engineering, Purdue University},
            addressline={610 Purdue Mall}, 
            city={West Lafayette},
            postcode={47907}, 
            state={IN},
            country={USA}}

\begin{abstract}

   We present PO-CKAN, a physics-informed deep operator framework based on Chunkwise Rational Kolmogorov–Arnold Networks (CKANs), for approximating the solution operators of partial differential equations. This framework integrates CKAN sub-networks within the Deep Operator Network (DeepONet) architecture, enabling physically consistent and computationally efficient operator learning. A physics-informed residual loss ensures that the learned operator adheres to governing PDE constraints. By embedding PDE-residual losses, PO-CKAN generalizes efficiently across varying parameters, boundary conditions, and initial states. Validated on challenging benchmark problems, PO-CKAN consistently achieves high-fidelity solutions and outperforms PI-DeepONet baselines. PO-CKAN adopts a DeepONet-style branch-trunk architecture with CKAN modules, and enforces physical consistency via a PDE residual (PINN-style) loss. On Burgers' equation with $\nu=0.01$, PO-CKAN reduces the mean relative $L^2$ error by approximately 48\% compared to PI-DeepONet, and achieves competitive accuracy on the Eikonal and diffusion-reaction benchmarks.These results highlight PO-CKAN as a scalable and interpretable framework for physics-consistent operator learning
\end{abstract}




\end{frontmatter}

\section{Introduction}

The Physics-Informed DeepONet (PI-DeepONet) \cite{lin_operator_2023} is an emerging machine learning framework designed to efficiently solve entire families of parameterized partial differential equations (PDEs) \cite{lucia_reduced-order_2004, mann_dynamic_2016, han_solving_2018, mezic_spectral_2005, psichogios_hybrid_1992, lagaris_artificial_1998}. Unlike standard physics-informed neural networks (PINNs), which typically solve individual PDE instances,\cite{raissi_physics-informed_2019}, PI-DeepONet approximates a universal functional mapping from an input function to its corresponding solution function.~\cite{park_deepsdf_2019, chen_universal_1995, back_universal_2002, raissi_physics-informed_2019, sun_surrogate_2020,zhu_physics-constrained_2019,karumuri_simulator-free_2020,sirignano_dgm_2018}. It achieves this through a distinctive “branch–trunk” architecture and ensures physical consistency by incorporating governing-equation residuals directly into the loss function. Its key advantage is the “train-once, predict-many” paradigm, which drastically reduces computational cost in settings requiring repeated PDE evaluations.

A continuing research direction seeks to enhance the expressive power and generalization capability of PI-DeepONet by replacing its multilayer-perceptron (MLP) submodules with more advanced operator-learning architectures—an approach that has already yielded promising results. \cite{cai_deepmmnet_2021, lin_operator_2023, lu_deeponet_2021}.

Kolmogorov-Arnold Networks (KANs) \cite{liu_kan_2024, kiamari_gkan_2024, samadi_smooth_2024, liu_initial_2024} have recently emerged as compelling alternatives to conventional MLPs. \cite{hornik_multilayer_1989} and are thus well-suited for this application. Grounded in the Kolmogorov–Arnold Representation Theorem, KANs differ fundamentally by placing learnable activation functions on the network edges rather than using fixed nonlinearities on the nodes.\cite{n_representations_1957, apicella_survey_2021, trentin_networks_2001} This architectural distinction leads to enhanced interpretability and robustness, particularly in problems with noisy data or continual-learning requirement \cite{yu_kan_2024, bozorgasl_wav-kan_2024,genet_tkan_2025,liu_ikan_2024, vaca-rubio_kolmogorov-arnold_2024, herbozo_contreras_kan-eeg_2025, peng_predictive_2024}.

Empirical studies have demonstrated that KANs can substantially outperform traditional MLPs in solving differential equations. For instance, KANs have been integrated into PINNs to solve the Poisson equation \cite{zhai_deep_2023}, and the KAN-based operator model DeepOKAN\cite{abueidda_deepokan_2024}, has been applied to wave-propagation and elasticity problems, achieving significant performance gains. However, despite their superior expressiveness, standard KANs suffer from a severe scalability bottleneck: the number of learnable parameters grows quadratically with network width, rendering naive integration into DeepONet computationally impractical for large-scale operator learning tasks \cite{cheon_demonstrating_2024, bodner_convolutional_2025, cheon_kolmogorov-arnold_2024}. This gap between theoretical potential and practical feasibility motivates the present work.

To address this challenge, we propose the \textbf{Chunk-wise Rational Kolmogorov–Arnold Network (CKAN)}—a parameter-efficient variant that employs chunk-wise parameter sharing and rational activation functions to drastically reduce computational overhead. Rational functions $R(x)=P(x)/Q(x)$ are well-known universal approximators capable of representing functions with singularities or sharp gradients \cite{babaei_solving_2024,boulle_rational_2020, telgarsky_neural_2017, leung_rational_1993}. By embedding such activations into a chunk-shared KAN structure, we obtain a scalable yet expressive operator-learning module.\cite{aghaei_rkan_2024, molina_pade_2020}. 

Integrating this efficient CKAN architecture into the physics-informed DeepONet framework yields the \textbf{Physics-Informed Deep Operator Chunk-rational KAN (PO-CKAN)}—a unified model that couples physical constraints with data-driven operator learning. The major contributions of this study are as follows:

\begin{itemize}
    \item We introduce an efficient rational activation function, providing a lightweight and numerically stable alternative to B-splines that preserves high expressive power while reducing computational cost.
    \item We design the Chunkwise Kolmogorov-Arnold Network (CKAN), a novel architecture that employs a chunk-wise parameter sharing mechanism to break the quadratic ($O(N^2)$) scaling bottleneck of traditional KANs.
    \item We develop the Physics-Informed Operator CKAN (PO-CKAN), integrating the CKAN within the DeepONet architecture to achieve the first scalable fusion of KAN-based expressivity and physics-informed operator learning.
    \item We demonstrate PO-CKAN’s accuracy and efficiency across four benchmark problems—including Burgers’, Eikonal, fractional, and diffusion–reaction PDEs—showing consistent improvements over PI-DeepONet baselines.
\end{itemize}

The structure of this paper is as follows. Section \ref{sec:Methology} introduces our proposed framework, the Physics-Informed Deep Operator Chunk-rational KAN (PO-CKAN). Subsequently, Section \ref{sec:numerical experiments} validates the framework's performance through a series of numerical experiments on benchmark problems, including the Eikonal Equation, Burgers' Equation, a Fractional PDE, and a Diffusion-reaction System. Finally, Section \ref{sec:conclusion} discusses our conclusions and outlines potential directions for future research.

\section{Methodology}
\label{sec:Methology}
\subsection{Preliminaries}

\subsubsection{Deep Operator Network (DeepONet)}
The Deep Operator Network (DeepONet) is a neural-operator framework designed to learn mappings between entire functional spaces rather than pointwise variable pairs.Unlike traditional surrogate models that approximate individual mappings between input and output variables, DeepONet directly learns the operator $\mathcal{G}: u(y)\rightarrow s(x)$ that transforms an input function $u(y)$ to its corresponding solution function $s(x)$. It employs a two-stream “branch–trunk” architecture consisting of specialized subnetworks:
\begin{itemize}
    \item \textbf{Branch Network}: Encodes the input function $u(y)$ into a $p$-dimensional latent representation, $b = [b_1, \dots, b_p] \in \mathbb{R}^p$.
    \item \textbf{Trunk Network}: Processes the spatio-temporal coordinates, $x$, to produce coordinate-dependent basis functions, $t(x) = [t_1(x), \dots, t_p(x)]$.
\end{itemize}
The final prediction is expressed as the inner product between branch and trunk features:
\begin{equation}
G(u)(x) \approx \sum_{k=1}^{p} b_k \cdot t_k(x) = \langle b, t(x) \rangle
\label{eq:deeponet}
\end{equation}

This decomposition decouples functional input encoding from spatial representation, yielding a flexible and memory-efficient structure for high-dimensional operator learning.

Once trained, a single DeepONet generalizes to new boundary conditions, forcing terms, or parameter settings without retraining, realizing the hallmark “train-once, predict-many” paradigm of operator learning.

Despite its strong performance, the expressive capacity of DeepONet is bounded by the representational limits of its MLP-based submodules, motivating the development of enhanced architectures—such as the proposed CKAN modules introduced in the next subsection—to better capture complex nonlinear functional dependencies.

\subsubsection{Kolmogorov–Arnold Network (KAN)}
Building upon the concepts introduced earlier, we now formally define the Kolmogorov–Arnold Network (KAN) architecture and its mathematical underpinnings.

The Kolmogorov–Arnold Network (KAN) is a neural architecture grounded in the Kolmogorov–Arnold Representation Theorem\cite{hecht-nielsen_kolmogorovs_1987}, which asserts that any multivariate continuous function can be expressed as a finite superposition of univariate functions and addition operations. This theorem provides a rigorous theoretical foundation for constructing neural networks capable of universal function approximation through one-dimensional transformations only.

For a specific smooth function $f: [0,1]^n \to \mathbb{R}$, the theorem guarantees that the following representation holds:
\begin{equation}
    f(\mathbf{x}) = f(x_1,\dots,x_n) = \sum_{q=1}^{2n+1} \Phi_q\left(\sum_{p=1}^n \phi_{q,p}(x_p)\right).
    \label{eq:KART}
\end{equation}
where both the inner functions $\phi_{q,p}: [0,1] \to \mathbb{R}$ and the outer functions $\Phi_q: \mathbb{R} \to \mathbb{R}$  are univariate. This formulation implies that addition is the only multivariate operation required, a remarkably elegant result. Yet, the original theorem’s constructive form involves highly irregular functions, so practical implementation demands smooth, learnable approximations.

To achieve this, the modern KAN parameterizes the univariate functions using smooth basis functions such as B-splines \cite{elfwing_sigmoid-weighted_2017, ruijters_gpu_2012, ruijters_efficient_2008, noauthor_chapter_nodate,boor_subroutine_1970}. Unlike conventional MLPs—which alternate between linear transformations and fixed nonlinear activations—KANs place learnable activation functions on the network edges instead of on the nodes.

A KAN with layer structure $[n_0, n_1, \dots, n_L]$. defines a sequence of function-matrix transformations:
\begin{equation*}
    x_{l+1,j} = \sum_{i=1}^{n_l} \phi_{l,j,i}(x_{l,i}), \quad j=1,\dots,n_{l+1}.
\end{equation*}
so that the overall network $\mathbf{\Phi}_l$ is:
\begin{equation*}
    \text{KAN}(\mathbf{x}) = (\mathbf{\Phi}_{L-1} \circ \mathbf{\Phi}_{L-2} \circ \cdots \circ \mathbf{\Phi}_0)(\mathbf{x}_0).
\end{equation*}
By contrast, an MLP uses linear weight matrices $\mathbf{W}$, and a single, fixed non-linear function, $\sigma$:
\begin{equation*}
    \text{MLP}(\mathbf{x}) = (\mathbf{W}_{L-1} \circ \sigma \circ \mathbf{W}_{L-2} \circ \sigma \circ \cdots \circ \mathbf{W}_0)(\mathbf{x}).
\end{equation*}
Thus, KANs integrate both the linear transformation and nonlinearity into learnable edge functions $\mathbf{\Phi}$, endowing them with higher expressivity and interpretability.

However, this expressive power comes at a cost. B-spline parameterization introduces a large number of coefficients and significant computational overhead, leading to quadratic parameter scaling with network width and hampering large-scale deployment.

Therefore, a more compact and computationally efficient variant is essential for practical operator-learning tasks—an issue directly addressed by the proposed Chunk-rational KAN (CKAN) architecture described next.

\subsection{The Proposed PO-CKAN Framework}
This section details the architecture and training objective of the Physics-Informed Deep Operator Chunk-rational KAN (PO-CKAN). We begin by describing the overall structure before providing a mathematical formulation of the novel CKAN layer. Finally, we define the components of the physics-informed loss function.

\subsubsection{Overall Architecture}

\begin{figure}[htbp]
    \centering
    \includegraphics[width=0.68\linewidth]{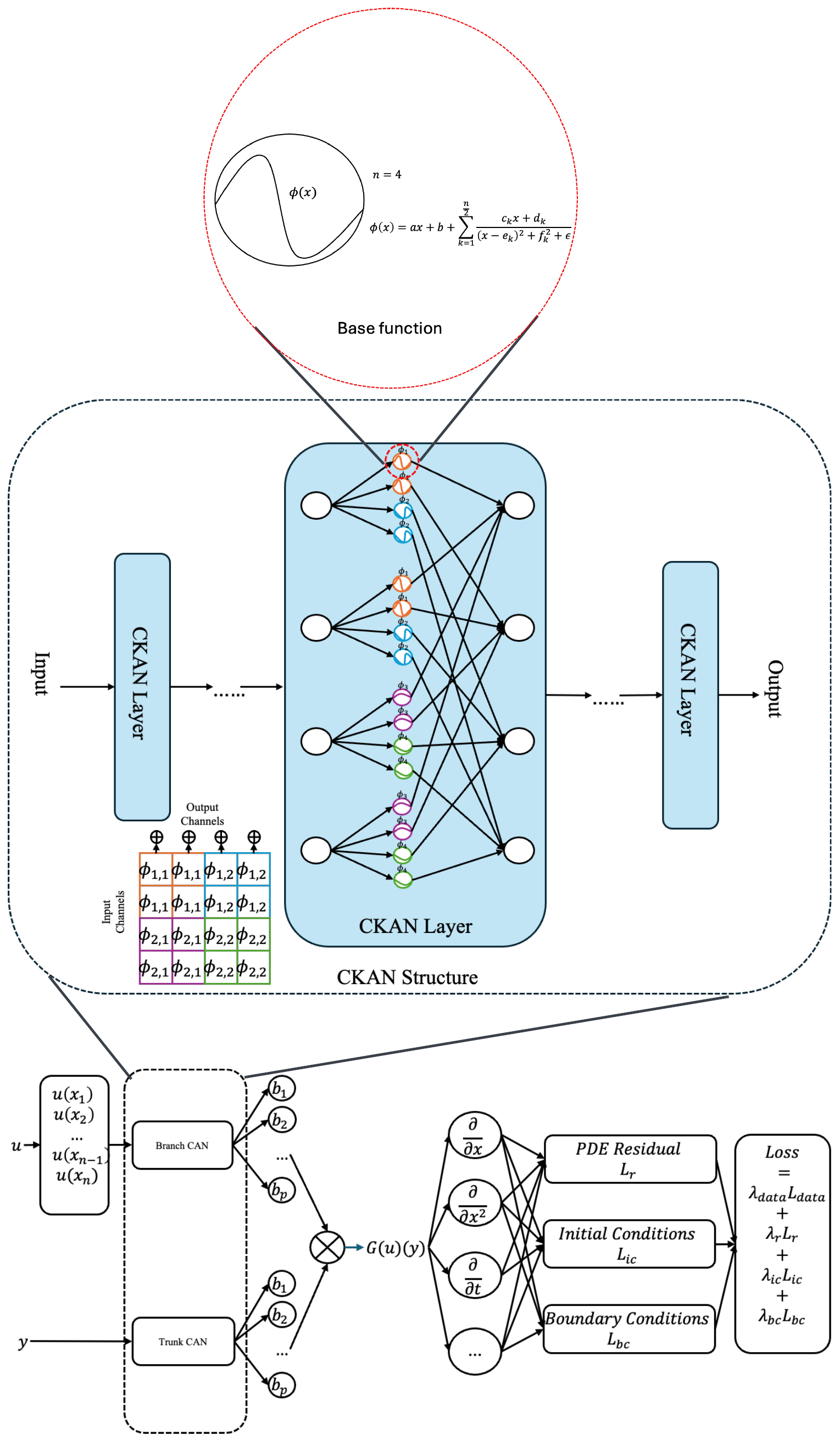} 
    \caption{This figure illustrates the PO-CKAN framework. The model adopts the DeepONet architecture to learn the solution operator mapping input functions to solution functions. The branch and trunk nets are both constructed from our novel CKAN layers. \textbf{Base Function (Top Panel):} The Enhanced Rational Unit (ERU) is used as the computationally efficient and numerically stable base function for all CKAN layers. \textbf{CKAN Layer (Middle Panel):} The CKAN layer is the core innovation. It reduces parameters through a chunk-wise sharing mechanism. Chunks of edges share a single base ERU but each edge retains an individual scalar weight. \textbf{Training Objective (Bottom Panel):} The loss is computed via automatic differentiation and comprises data ($\mathcal{L}_{\text{data}}$), initial condition ($\mathcal{L}_{\text{ic}}$), boundary condition ($\mathcal{L}_{\text{bc}}$), and PDE residual ($\mathcal{L}_{r}$) terms.}
    \label{fig:overall_architecture}
\end{figure}

The Physics-Informed Deep Operator Chunk-rational Kolmogorov–Arnold Network (PO-CKAN) inherits the canonical “branch–trunk” structure of DeepONet, as illustrated in~\ref{fig:overall_architecture}. he architecture comprises two complementary components: a branch network that processes the input function $u(y)$, and a trunk network that encodes the spatial–temporal coordinates $x$. Both subnetworks are constructed entirely from our newly developed Chunk-rational KAN (CKAN) layers, which replace the conventional MLP blocks used in standard DeepONet. 

This substitution enables the model to capture complex nonlinear functional dependencies while remaining computationally tractable. Each CKAN layer provides a learnable, parameter-efficient transformation that maintains the representational richness of Kolmogorov–Arnold Networks but mitigates their quadratic parameter growth through chunk-wise parameter sharing and rational activation functions.

The overall PO-CKAN framework thus unifies physics-based constraints, operator learning, and scalable architecture design. During training, the model minimizes a composite physics-informed loss function incorporating data fidelity, initial/boundary conditions, and PDE residual terms (as detailed in Section \ref{sec:pde loss}).

Collectively, these innovations enable PO-CKAN to deliver high-fidelity, physics-consistent operator approximations with markedly improved efficiency compared with PI-DeepONet baselines.
\subsubsection{CKAN Layer}
The Chunk-rational Kolmogorov–Arnold Network (CKAN) constitutes the core architectural innovation of PO-CKAN, addressing the long-standing scalability limitations of standard KANs.

Traditional KANs exhibit quadratic parameter growth with network width due to the use of edge-specific B-spline activations, leading to high memory cost and computational overhead.

The CKAN layer overcomes these bottlenecks through two complementary mechanisms:
\begin{itemize}
    \item Efficient and Stable Rational Activations
    \item Chunk-wise Parameter Sharing
\end{itemize}

\paragraph{\textbf{Efficient and Stable Rational Activations}}
To enhance expressiveness while maintaining numerical stability, CKAN replaces the conventional B-spline basis with smooth rational activation functions. \cite{leung_rational_1993, boulle_rational_2020}.

Rational forms are particularly adept at approximating functions that contain sharp gradients, discontinuities, or singularities.

Each edge function is parameterized as:
\begin{equation}
    \phi(x) = w F(x) = w \cdot \frac{P(x)}{Q(x)} = w \cdot \frac{a_0 + a_1 x + \cdots + a_m x^m}{b_0 + b_1 x + \cdots + b_n x^n},
    \label{eq:rational_phi_general}
\end{equation}
where the numerator $P(x)$ and denominator $Q(x)$ are polynomials of degrees $m$ and $n$, respectively.

To prevent numerical instability from zero denominators, CKAN adopts the Enhanced Rational Unit (ERU) \cite{avidan_era_2022}, , which guarantees a positive denominator and smooth behavior:
\begin{equation}
    F(x) = ax+b + \sum^{\frac{n}{2}}_{k=1}\frac{c_kx+d_k}{(x-e_k)^2+f_k^2+\epsilon},
    \label{eq:eru}
\end{equation}
where $\epsilon > 0$ ensures stability.

This design yields significantly lower FLOP counts than the traditional B-spline activation (about $10 \times$ reduction) while retaining high expressive power, as summarized in Table~\ref{tab:flops_comparison_booktabs}

Hence, ERUs enable CKAN layers to achieve efficient and numerically robust nonlinearity modeling within a physics-informed operator-learning context.

\begin{table}[h!]
\centering
\begin{tabular}{lcr}
\toprule
\textbf{Model} & \textbf{Setting} & \textbf{FLOPs} \\ 
\midrule
B-Spline & $G=3,\ K=3$ & 204 \\
Rational & $m=5,\ n=4$ & 46 \\
\textbf{ERU} & $n=4$ & \textbf{19} \\ 
\bottomrule
\end{tabular}
\caption{FLOP analysis of various function types. The Enhanced Rational Unit achieves roughly a $10.7\times$ reduction in FLOPs compared with the B-Spline formulation.}
\label{tab:flops_comparison_booktabs}
\end{table}

\paragraph{\textbf{Chunkwise Parameter Sharing}}
The second major improvement, chunk-wise parameter sharing, mitigates the $O(d_{\text{in}}d_\text{out})$ growth of activation functions in vanilla KANs. 

The core concept is to share the base parameters of a rational function across a chunk of connections rather than learning a unique function for each input-output pair:
\begin{enumerate}
    \item \textbf{Chunking}: We divide the $d_{\text{in}}$ input channels and $d_{\text{out}}$ output channels into $c \times c$ chunks.
    \item \textbf{Parameter Sharing}: Within each chunk $(m,n)$, all edges share a \textbf{single} base rational function, $F_{m,n}(x)$, in the form of Eq.~\ref{eq:eru}.
    \item \textbf{Edge-specific Weights}: To preserve expressive power, we retain a unique, learnable scalar weight $w_{ij}$ for each individual edge $(i, j)$ in the layer.
\end{enumerate}

\begin{figure}[htbp]
    \centering
    \includegraphics[width=0.9\linewidth]{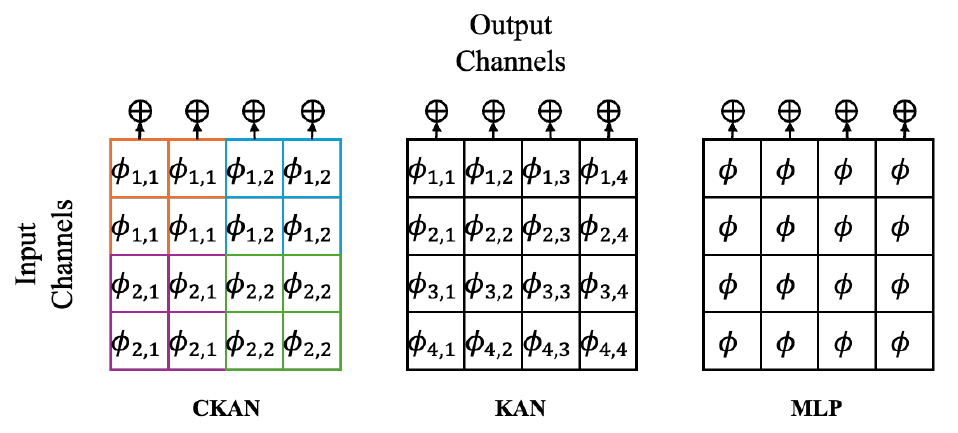} 
    \caption{A comparison between our CKAN (configured with a 2×2 chunk) and standard KAN and MLP models. Unlike the conventional KAN, which assigns a distinct function to every input–output connection, CKAN employs one shared base function for each chunk of edges.}
    \label{fig:Comparison_of_Models}
\end{figure}

Formally, for a CKAN layer mapping $x \in \mathbb{R}^d_{\text{in}}$ to $\mathbb{R}^d_{\text{out}}$

\begin{equation}
\text{CKAN}(x)_j = \sum_{i=1}^{d_{\text{in}}} w_{ij} \cdot F_{\lfloor i / d_{\text{in},c} \rfloor, \lfloor j / d_{\text{out},c} \rfloor}(x_i),
\end{equation}
where $d_{\text{in},c} = d_{\text{in}}/c$ and $d_{\text{out},c} = d_{\text{out}}/c$ are the chunk sizes for the input and output dimensions, respectively, and $(\lfloor i / d_{\text{in},c} \rfloor, \lfloor j / d_{\text{out}, c} \rfloor)$ identifies the chunk index.

In a standard KAN, $d_{\text{in}} \times d_{\text{out}}$ distinct activations must be evaluated; By CKAN reduces this to only $c \times c$. Consequently, both parameter count and FLOP complexity decrease by several orders of magnitude while maintaining nearly identical inference latency in practice.

Consequently, both parameter count and FLOP complexity decrease by several orders of magnitude while maintaining nearly identical inference latency in practice.

Illustratively, Figure~\ref{fig:Comparison_of_Models} compares KAN, CKAN, and MLP layers: while a standard KAN uses a unique activation per connection, CKAN reuses shared ERU bases within each chunk, dramatically reducing computational overhead.

Table~\ref{tab:param_comparison_original} further quantifies this gain, showing that CKAN retains MLP-like parameter counts while achieving KAN-level functional flexibility.

\begin{table}[htbp]
\centering
\resizebox{\linewidth}{!}{%
\begin{tabular}{@{}lll@{}}
\toprule
\textbf{Architecture} & \textbf{Parameter Count} & \textbf{Estimated FLOPs} \\
\midrule
MLP & $d_\text{in} \, d_\text{out} + d_\text{out}$ & $d_\text{out} \cdot \text{Func FLOPs} + 2 \, d_\text{in} d_\text{out}$ \\[1.5mm]
KAN & $d_\text{in} \, d_\text{out} \, (G + K + 3) + d_\text{out}$ & $d_\text{in} \cdot \text{Func FLOPs} + d_\text{in} d_\text{out} \,[\,9 K (G + 1.5 K) + 2 G - 2.5 K + 3\,]$ \\[1.5mm]
\textbf{CKAN} & $d_\text{in} \, d_\text{out} + d_\text{out} + (2 n + 2) c^2$ & $ d_\text{in}\cdot \,(4.5 n + 1) \, c + 2 \, d_\text{in} d_\text{out}$ \\
\bottomrule
\end{tabular}
}
\caption{Comparison of parameter complexity and computational cost across different neural architectures. Here, \textit{Func FLOPs} denotes the floating-point operations required to evaluate the non-linear activation function. In KAN, $K$ specifies the base function degree and $G$ the number of grid points. CKAN employs rational functions of order $n$ with a $c \times c$ chunk configuration, resulting in a modest fixed parameter overhead relative to a standard MLP, whereas KAN scales linearly with $(G+K+3)$.}
\label{tab:param_comparison_original}
\end{table}

\subsubsection{Physics-Informed Training Objective}
\label{sec:pde loss}
A major challenge in operator learning is that purely data-driven models often yield solutions that fit the training data but violate underlying physical laws. To overcome this limitation, PO-CKAN embeds the governing partial differential equations (PDEs) directly into the optimization objective, ensuring that the learned operator satisfies both empirical data and physical constraints.

The network parameters $\theta$ are trained by minimizing a composite physics-informed loss function:
\begin{equation}
    \mathcal{L}(\theta) = 
    \lambda_{\mathrm{data}}\,\mathcal{L}_{\mathrm{data}}(\theta) +
    \lambda_{\mathrm{ic}}\,\mathcal{L}_{\mathrm{ic}}(\theta) +
    \lambda_{\mathrm{bc}}\,\mathcal{L}_{\mathrm{bc}}(\theta) +
    \lambda_{\mathrm{r}}\,\mathcal{L}_{\mathrm{r}}(\theta),
\end{equation}
where the coefficients $\lambda_{\ast} \geq 0$ balance the contribution of data, initial-condition, boundary-condition, and residual losses, respectively.

\paragraph{\textbf{Data loss}}
The data-fidelity term enforces agreement between model predictions and available reference solutions:
\begin{equation}
    \mathcal{L}_{\mathrm{data}}(\theta) = \frac{1}{N_{d}} 
    \sum_{i=1}^{N_{d}} \left\| G_{\theta}(u_i)(y_i) - s_i(y_i) \right\|_2^2.
\end{equation}

This component constrains the network to reproduce high-fidelity simulation or experimental results whenever labeled data are available.

\paragraph{\textbf{Initial and Boundary Condition Losses}}
The initial-condition loss ensures consistency with the PDE’s prescribed initial state $s(y,0) = s_0(y)$:
\begin{equation}
    \mathcal{L}_{\mathrm{ic}}(\theta) = \frac{1}{N_{\mathrm{ic}}} 
    \sum_{j=1}^{N_{\mathrm{ic}}} \left\| G_{\theta}(u_j)(y_j^{\mathrm{ic}}, 0) - s_0(y_j^{\mathrm{ic}}) \right\|_2^2,
\end{equation}
while the boundary-condition loss enforces either Dirichlet or Neumann constraints along the boundary domain:
\begin{equation}
    \mathcal{L}_{\mathrm{bc}}(\theta) = \frac{1}{N_{\mathrm{bc}}} 
    \sum_{j=1}^{N_{\mathrm{bc}}} \left\| G_{\theta}(u_j)(y_j^{\mathrm{bc}}) - s_{\mathrm{bc}}(y_j^{\mathrm{bc}}) \right\|_2^2.
\end{equation}

Together, these two terms guarantee that the learned operator adheres to physical and geometric boundary specifications.

\paragraph{\textbf{PDE residual loss}}
The residual term penalizes violations of the governing equation $\mathcal{R}(u, s) = 0$ evaluated over a set of collocation points $\{y_j^{\mathrm{phys}}\}$:
\begin{equation}
    \mathcal{L}_{\mathrm{r}}(\theta) = \frac{1}{N_{p}} 
    \sum_{j=1}^{N_{p}} \left\| \mathcal{R}\big(u_j, G_{\theta}(u_j)\big)\!\left(y_j^{\mathrm{phys}}\right) \right\|_2^2.
\end{equation}

All spatial and temporal derivatives required for evaluating the residual are computed via automatic differentiation, ensuring analytical accuracy and stability.

By jointly minimizing these four components, PO-CKAN learns an operator that is simultaneously data-consistent and physics-compliant.
This hybrid loss framework allows the network to generalize effectively across unseen parameter regimes, boundary conditions, and initial states, while inherently respecting the underlying PDE structure—a defining feature that differentiates PO-CKAN from purely data-driven operator networks.

\section{Numerical Experiments}
\label{sec:numerical experiments}

\subsection{Burgers' Equation}
We first evaluate PO-CKAN on the one-dimensional viscous Burgers’ equation, a canonical benchmark for nonlinear wave propagation and shock formation:
\begin{equation}
\begin{cases}\label{burger_eq_final}
    \frac{\partial s}{\partial t} + s\, \frac{\partial s}{\partial x} - \nu\, \frac{\partial^2 s}{\partial x^2} = 0, & \text{for } x \in (0,1),\ t \in (0,1),\\
    s(x,0) = u(x), & \text{for } x \in (0,1),
\end{cases}
\end{equation}
subject to periodic boundary conditions.

he learning objective is to approximate the operator mapping the initial condition $u(x)$ to the corresponding solution field $s(x,t)$. Initial conditions are sampled from a Gaussian Random Field (GRF), $u \sim \mathcal{N}\left(0, 25^2(-\Delta + 5^2 I)^{-4} \right)$, ensuring a diverse range of smooth and oscillatory inputs.

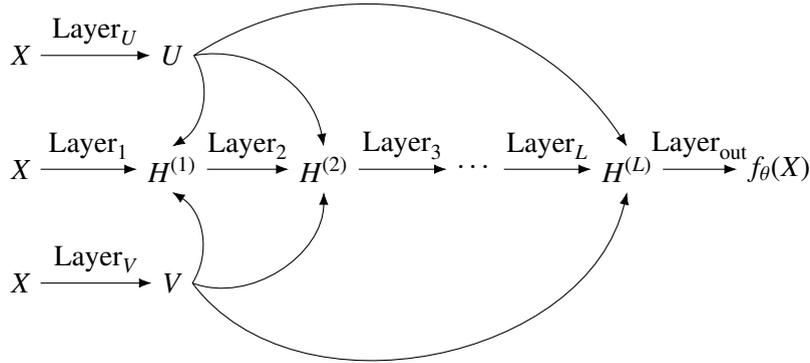
\begin{figure}[ht!] 
\centering 
\begin{tikzpicture}[node distance=1.5cm, auto]
    \node (X1) at (0,0) {$X$};
    \node (U) at (2,0) {$U$};
    
    \node (X2) at (0,-1.5) {$X$};
    \node (H1) at (2,-1.5) {$H^{(1)}$};
    \node (H2) at (4,-1.5) {$H^{(2)}$};
    \node (dots) at (6,-1.5) {$\cdots$};
    \node (HL) at (8,-1.5) {$H^{(L)}$};
    
    \node (X3) at (0,-3) {$X$};
    \node (V) at (2,-3) {$V$};
    
    \node (f) at (10,-1.5) {$f_{\theta}(X)$};

    \draw[-{Latex}] (X1) -- node[above] {$\text{Layer}_U$} (U);
    
    \draw[-{Latex}] (X2) -- node[above] {$\text{Layer}_1$} (H1);
    \draw[-{Latex}] (H1) -- node[above] {$\text{Layer}_2$} (H2);
    \draw[-{Latex}] (H2) -- node[above] {$\text{Layer}_3$} (dots);
    \draw[-{Latex}] (dots) -- node[above] {$\text{Layer}_L$} (HL);
    
    \draw[-{Latex}] (HL) -- node[above] {$\text{Layer}_{\text{out}}$} (f);
    
    \draw[-{Latex}] (X3) -- node[above] {$\text{Layer}_V$} (V);
    
    \draw[-{Latex}, bend left=45] (U.east) to (H1.north);
    \draw[-{Latex}, bend left=45] (U.east) to (H2.north);
    \draw[-{Latex}, bend left=45] (U.east) to (HL.north);
    
    \draw[-{Latex}, bend right=45] (V.east) to (H1.south);
    \draw[-{Latex}, bend right=55] (V.east) to (H2.south);
    \draw[-{Latex}, bend right=65] (V.east) to (HL.south);
\end{tikzpicture}
\caption{Modified network architecture}
\label{fig:modified network}
\end{figure}

Following the benchmark dataset of~\cite{wang_learning_2021}, we generate 2,000 GRF samples, using 1,500 for training and 500 for testing.

Ground-truth solutions are obtained via a spectral Chebfun solver~\cite{cox_exponential_2002}, with 101 temporal snapshots saved per solution. 

\paragraph{\textbf{Network Architecture and Training Setup}}

In our experiments, following~\cite{wang_understanding_2020}, we adopt the modified network as the base architecture, which incorporates residual connections and gate-controlled mechanisms ~\ref{fig:modified network}. The modified network exhibits a distinctive forward propagation mechanism \cite{wang_understanding_2020} as follows

\begin{equation}
\begin{aligned}
U &= \text{Layer}_U(X), \quad V = \text{Layer}_V(X), \\
H^{(1)} &= \text{Layer}_1(X), \\
Z^{(k)} &= \text{Layer}_k(H^{(k)}), \quad k = 1, \dots, L, \\
H^{(k+1)} &= (1 - Z^{(k)}) \odot U + Z^{(k)} \odot V, \quad k = 1, \dots, L, \\
f_\theta(x) &= \text{Layer}_\text{out}(H^{(L)}).
\end{aligned}
\end{equation}

Both the branch and trunk networks of PO-CKAN employ a four-layer CKAN (rational degree \(n=4\), chunk number \(c=1\times 1\)), with 100 units per layer. 

For comparison, the baseline PI-DeepONet uses an identical architecture but replaces CKAN layers with tanh-activated MLPs. 

Training uses the Adam optimizer for 100,000 iterations, guided purely by the physics-informed loss that enforces the initial, boundary, and residual constraints:

\begin{equation}
\mathcal{L}(\theta) = \lambda_{\textbf{ic}}\mathcal{L}_{\text{ic}}(\theta) + \lambda_{\textbf{bc}}\mathcal{L}_{\text{bc}}(\theta) + \lambda_{\textbf{r}}\mathcal{L}_{\text{r}}(\theta), \, (\lambda_{\textbf{ic}},\lambda_{\textbf{bc}}, \lambda_{\textbf{r}})=(50, 1, 1)
\label{eq:burger_loss_function_final}
\end{equation}

Experiments are performed for viscosity coefficients, $\nu \in \{0.05, 0.03, 0.01\}$, smaller $\nu$ produces sharper gradients and thus constitutes a more challenging nonlinear regime. 

\begin{figure}[H]
    \centering
    \includegraphics[width=1\linewidth]{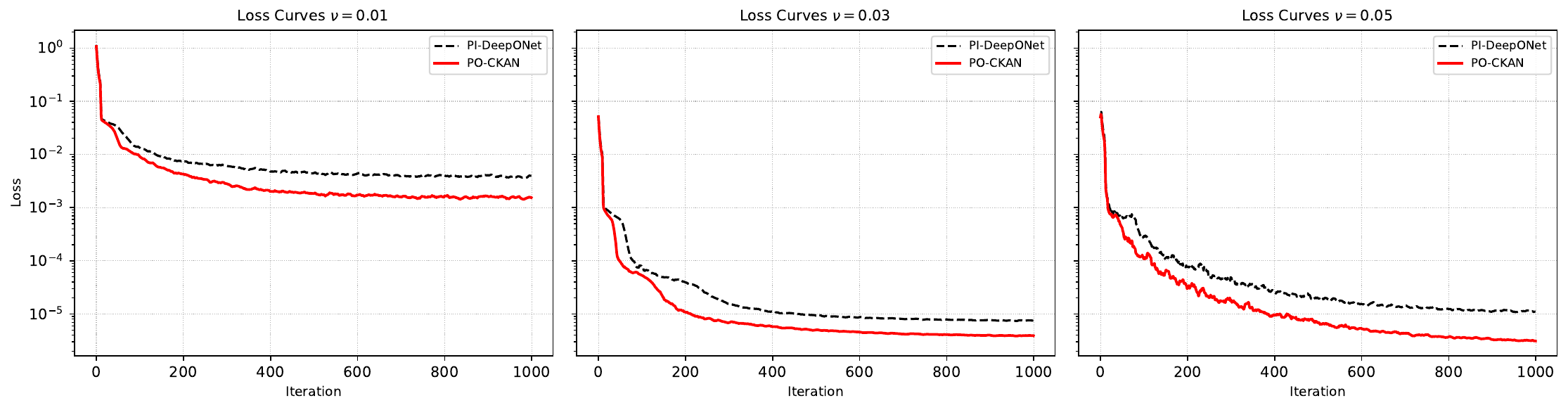}
    \caption{PO-CKAN vs. PI-DeepONet:loss at different viscosity coefficients $\nu$. ($\nu$ = 0.01, 0.03, 0.05)}
    \label{burger_loss}
\end{figure}

\begin{figure}[H]
  \centering
  \begin{subfigure}{\linewidth}
    \centering
    \includegraphics[width=0.8\linewidth]{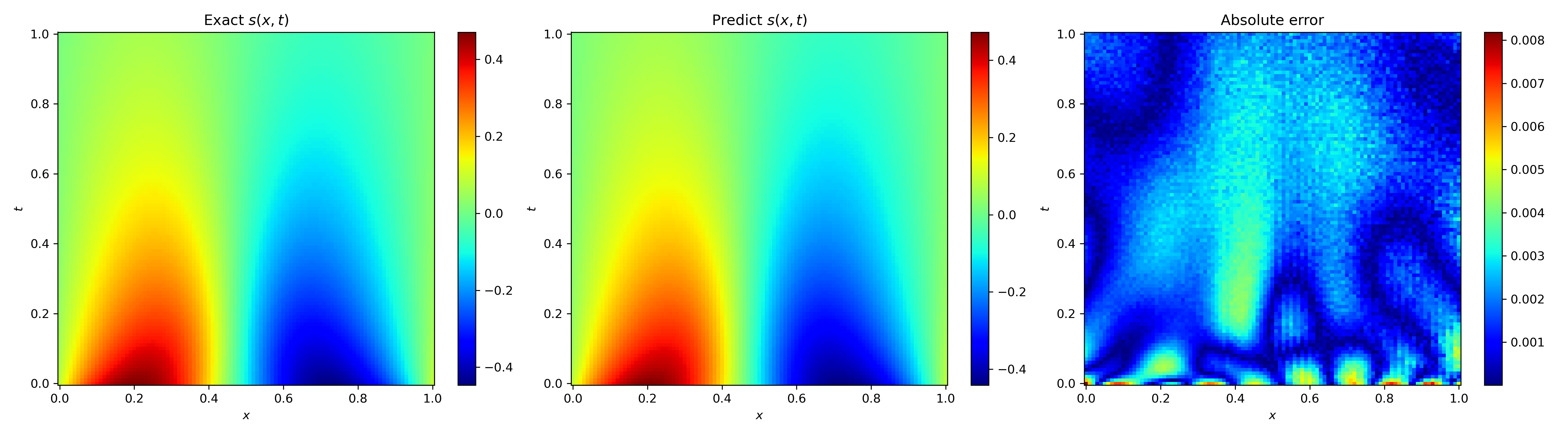}
    \caption*{PI-DeepONet}%
  \end{subfigure}
  \vspace{-0.3em}  
  \begin{subfigure}{\linewidth}
    \centering
    \includegraphics[width=0.8\linewidth]{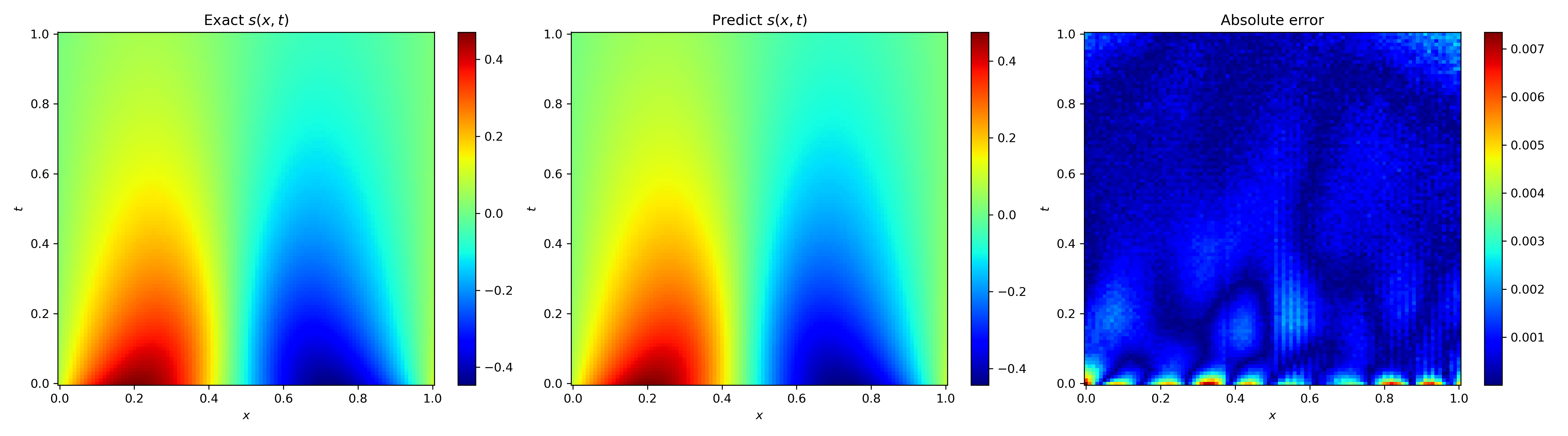}
    \caption*{PO-CKAN}
  \end{subfigure}
  \vspace{-0.5em} 
  \caption{Visualization of PI-DeepONet and PO-CKAN outputs for Burgers' equation ($\nu=0.05$). Shown, from left to right, are the exact solution, the network approximation, and the absolute error.
}
  \label{fig:burger_005}
\end{figure}

\begin{figure}[H]
  \centering
  \begin{subfigure}{\linewidth}
    \centering
    \includegraphics[width=0.8\linewidth]{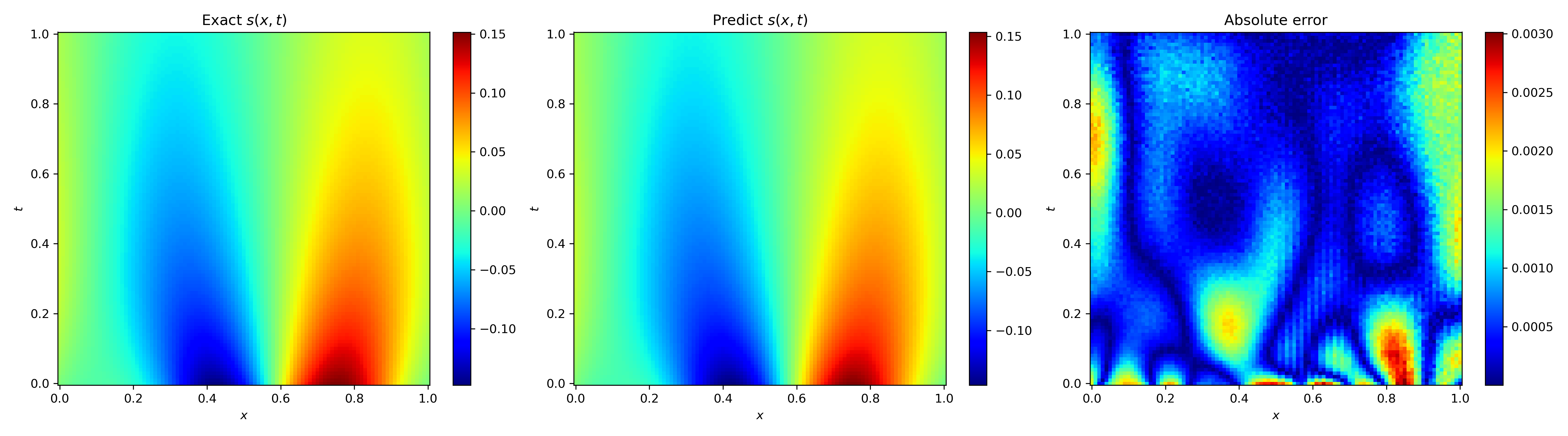}
    \caption*{PI-DeepONet}%
  \end{subfigure}
  \vspace{-0.3em}  
  \begin{subfigure}{\linewidth}
    \centering
    \includegraphics[width=0.8\linewidth]{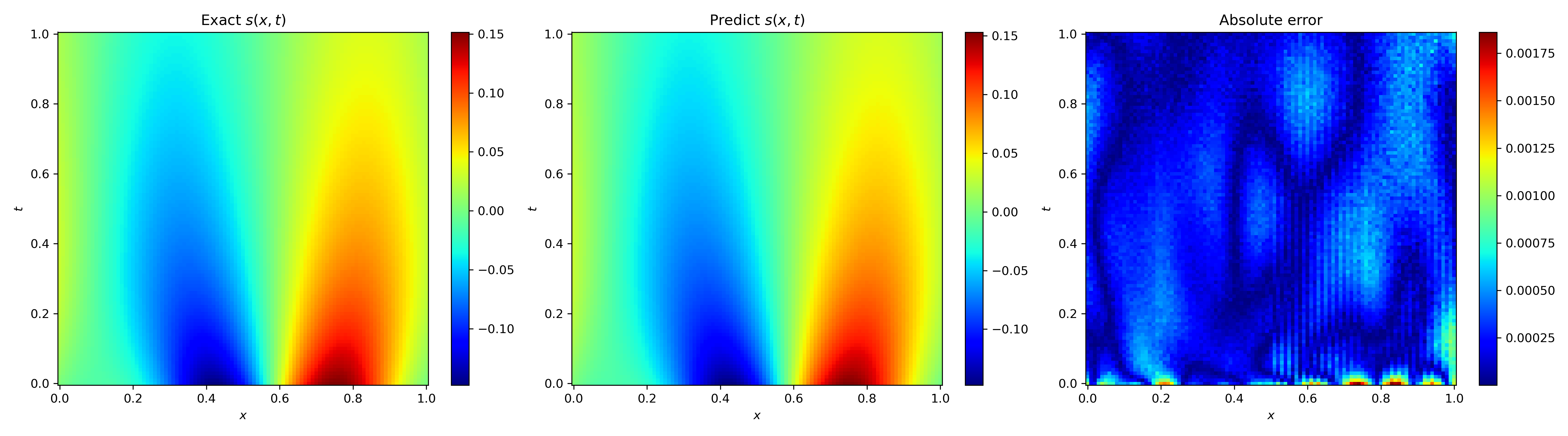}
    \caption*{PO-CKAN}
  \end{subfigure}
  \vspace{-0.5em} 
  \caption{Visualization of PI-DeepONet and PO-CKAN outputs for Burgers' equation ($\nu=0.03$). Shown, from left to right, are the exact solution, the network approximation, and the absolute error.
}
  \label{fig:burger_003}
\end{figure}

\begin{figure}[H]
  \centering
  \begin{subfigure}{\linewidth}
    \centering
    \includegraphics[width=0.8\linewidth]{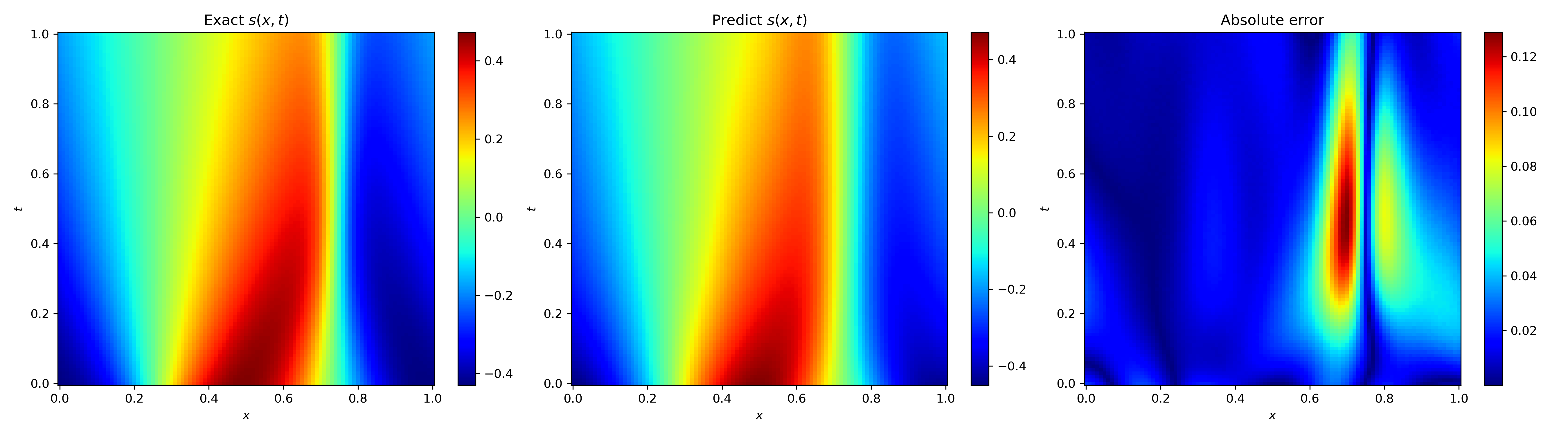}
    \caption*{PI-DeepONet}%
  \end{subfigure}
  \vspace{-0.3em}  
  \begin{subfigure}{\linewidth}
    \centering
    \includegraphics[width=0.8\linewidth]{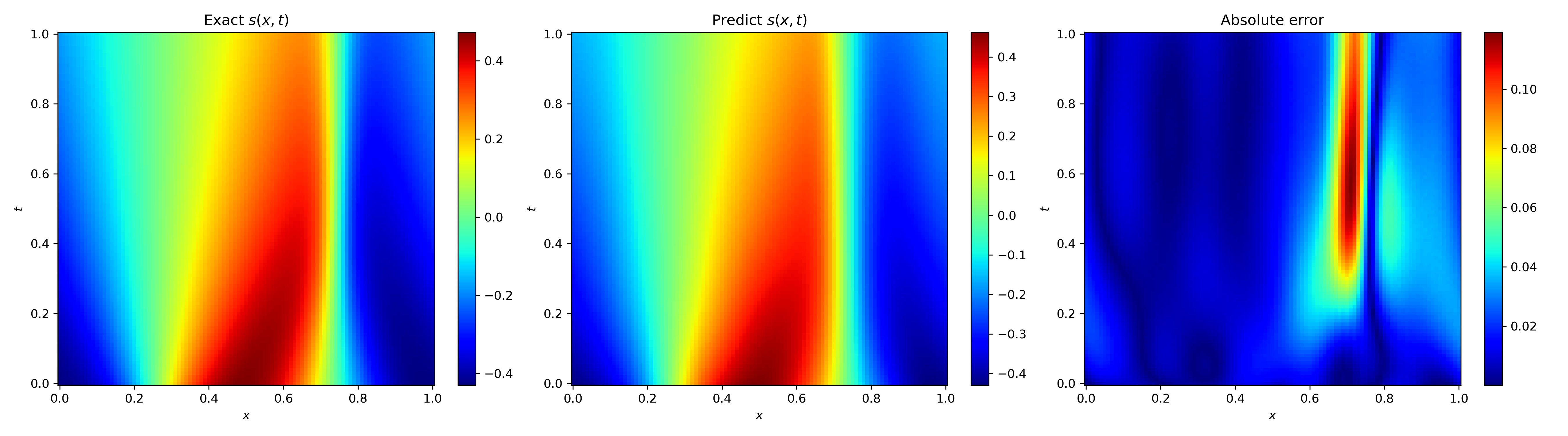}
    \caption*{PO-CKAN}
  \end{subfigure}
  \vspace{-0.5em} 
  \caption{Visualization of PI-DeepONet and PO-CKAN outputs for Burgers' equation ($\nu=0.01$). Shown, from left to right, are the exact solution, the network approximation, and the absolute error.
}
  \label{fig:burger_001}
\end{figure}

\paragraph{\textbf{Results and Analysis}}
The results, presented in Figures \ref{burger_loss},\ref{fig:burger_005},\ref{fig:burger_003},\ref{fig:burger_001} and Tables \ref{tab:burger-errors},\ref{tab:burger-losses}, summarize the comparative performance.
PO-CKAN consistently outperforms PI-DeepONet across all viscosity settings. Training-loss curves in Figure~\ref{burger_loss} show faster convergence and lower asymptotic error for PO-CKAN.
Qualitative comparisons (Figures~\ref{fig:burger_005},\ref{fig:burger_003},\ref{fig:burger_001}) demonstrate that PO-CKAN more accurately resolves steep shock fronts, especially for $\nu = 0.01$, while the baseline suffers from visible diffusion errors. Point-wise error maps confirm these observations quantitatively.

As listed in Table~\ref{tab:burger-errors}, PO-CKAN reduces the mean relative $L_2$ error by approximately 48\% for $\nu=0.01$; the final loss values (Table~\ref{tab:burger-losses}) likewise show more than a twofold improvement.

These results underscore that the CKAN design enhances the stability and representational capacity of physics-informed operator networks, particularly in regimes exhibiting strong nonlinearity and multiscale features.

\begin{table}[H]
    \centering
    \begin{tabular}{c|c|c|c}
      \hline
      \diagbox{Model}{viscosity}
      & $\nu = 0.05$ 
      & $\nu = 0.03$ 
      & $\nu = 0.01$ \\
      \hline
      PI-DeepONet    
      & $1.22\times 10^{-2}$ 
      & $1.38\times 10^{-2}$ 
      & $6.23\times 10^{-2}$ \\
      \hline
      \textbf{PO-CKAN}    
      & $\mathbf{6.93\times 10^{-3}}$
      & $\mathbf{7.03\times 10^{-3}}$ 
      & $\mathbf{3.21\times 10^{-2}}$ \\
      \hline      
    \end{tabular}
    \caption{Mean relative $L^2$ errors of PI-DeepONet and PO-CKAN for the Burgers' equation with different viscosity coefficients~$\nu$.}
    \label{tab:burger-errors}
\end{table}

\begin{table}[H]
    \centering
    \begin{tabular}{c|c|c|c}
      \hline
      \diagbox{Model}{viscosity}
      & $\nu = 0.05$ 
      & $\nu = 0.03$ 
      & $\nu = 0.01$ \\
      \hline
      PI-DeepONet    
      & $3.41\times 10^{-4}$ 
      & $5.59\times 10^{-4}$ 
      & $3.66\times 10^{-3}$ \\
      \hline
      \textbf{PO-CKAN}    
      & $\mathbf{1.33\times 10^{-4}}$ 
      & $\mathbf{1.77\times 10^{-4}}$ 
      & $\mathbf{{1.41\times 10^{-3}}}$ \\
      \hline      
    \end{tabular}
    \caption{Final losses of PI-DeepONet and PO-CKAN for the Burgers' equation with different viscosity coefficients~$\nu$.}
    \label{tab:burger-losses}
\end{table}

\subsection{Eikonal Equation}

We next evaluate PO-CKAN on a geometric operator-learning problem governed by the two-dimensional Eikonal equation, which arises in computing Signed Distance Functions (SDFs) used extensively in computer vision, robotics, and computational geometry~\cite{park_deepsdf_2019}. Formally, the problem is defined on a domain $\Omega \subset \mathbb{R}^2$ as
\begin{equation}
\label{eikonal_equation}
\|\nabla s(\mathbf{x})\|_2 = 1, \quad \mathbf{x} \in \Omega, \quad s(\mathbf{x}) = 0 \quad \text{if}\ \mathbf{x} \in \partial \Omega,
\end{equation}
where $\mathbf{x} = (x, y)$ and the boundary $\partial \Omega$ is a closed curve $\Gamma$. The solution $s(\mathbf{x})$ represents the minimal distance from any point $\mathbf{x}$ in the domain to the boundary $\Gamma$. Our objective is to approximate the solution operator
\[
G: \Gamma \mapsto s(\mathbf{x}),
\]
which maps a given boundary shape to its corresponding SDF.

\paragraph{\textbf{Problem Setup and Data Generation}}
For benchmarking, we consider circular boundaries centered at the origin, for which an analytical solution exists: the SDF for a circle $\Gamma$ of radius $r$ is given by: 
\begin{equation}
    s(x, y) = \sqrt{x^2 + y^2} - r
\end{equation}
Following~\cite{wang_learning_2021}, We generate 1,000 training examples by sampling $r$ from $U(0.5, 1.5)$. Each circle $\Gamma^{(i)}$ discretized into 100 points, and model predictions are evaluated over a square domain $D = [-2, 2] \times [-2, 2]$.

\paragraph{\textbf{Network Architecture and Training Protocol}}
The branch and trunk networks of PO-CKAN each employ a four-layer CKAN (rational degree $n=4$, chunk configuration $2 \times 2$) with 50 units per layer. The baseline PI-DeepONet uses the same topology but with standard MLP layers and tanh activations. To highlight the benefits of our approach, we deliberately use a shallow baseline.

Both models are trained for 80,000 iterations using the Adam optimizer, driven solely by a physics-informed loss enforcing boundary and residual terms:
\begin{equation}\label{eikonal_loss}
    \mathcal{L}(\theta) = \lambda_{\text{bc}}\mathcal{L}_{\text{bc}}(\theta) + \lambda_{\text{r}}\mathcal{L}_{\text{r}}(\theta), \, (\lambda_{\text{bc}},\lambda_{\text{r}})=(1,1)
\end{equation}
The boundary-condition loss penalizes deviation of predicted SDF values from zero along the boundary,
while the residual loss enforces the PDE constraint at $Q=1,000$ collocation points $\mathbf{x} \in \Omega$. In this notation, $\Gamma^{(i)}$ denotes the $i$-th input boundary curve, $\mathbf{x}$ represents a point in the computational domain, and $Q$ is the number of collocation points. The core residual is computed as:
\begin{equation}
    R_\theta^{(i)}(\mathbf{x}) = \left(\frac{\partial G_\theta(\Gamma^{(i)})}{\partial x}\right)^2 + \left(\frac{\partial G_\theta(\Gamma^{(i)})}{\partial y}\right)^2 - 1.
\end{equation}

\begin{figure}[htbp]
    \centering
    \includegraphics[width=1\linewidth]{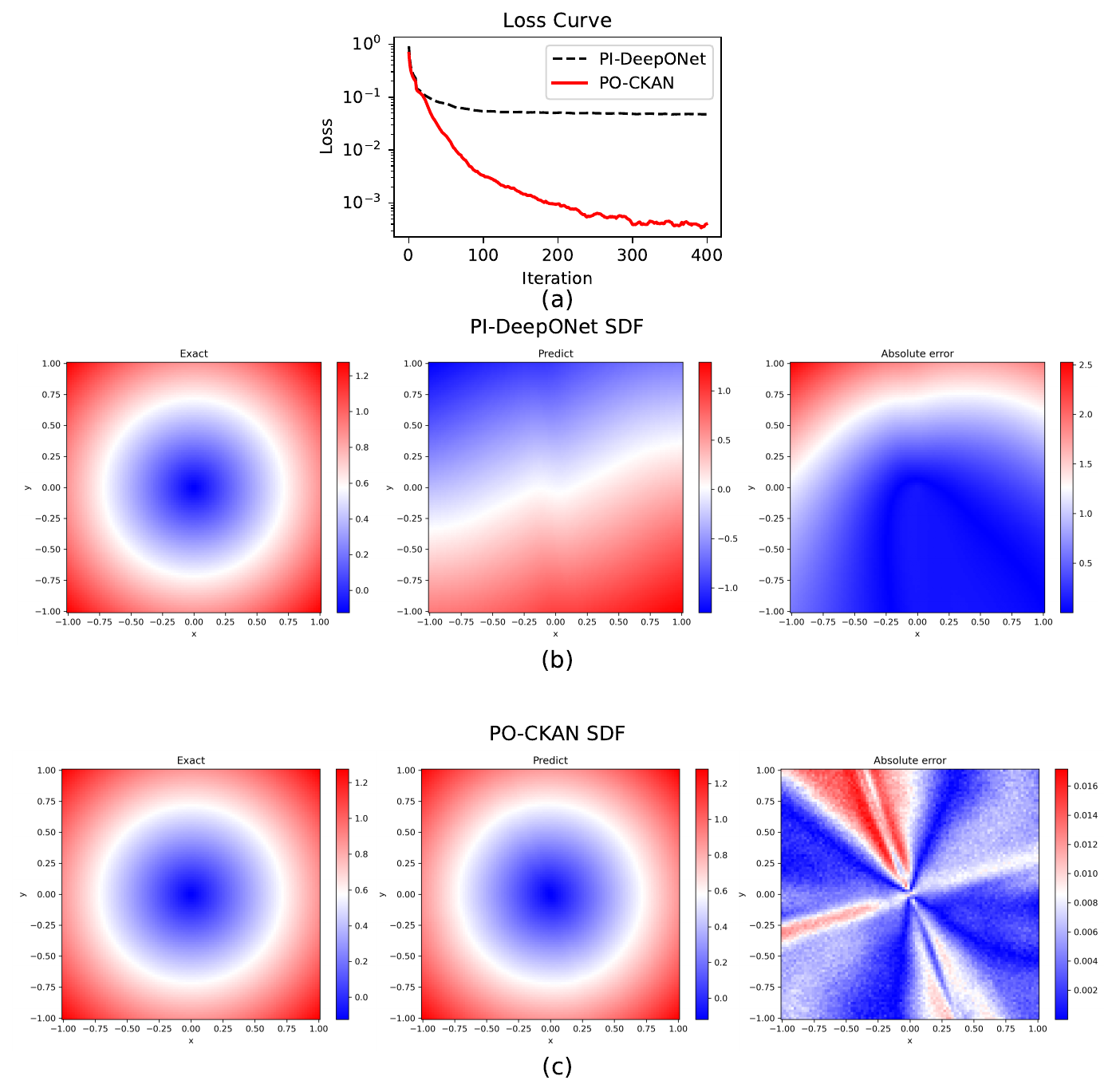}
    \caption{
    \textbf{Comprehensive performance comparison between PO-CKAN and PI-DeepONet for learning a circular Signed Distance Function (SDF).}
    Figure (a) shows the test loss curves, where PO-CKAN converges to a loss approximately two orders of magnitude lower than PI-DeepONet ($\sim 10^{-3}$ vs. $\sim 10^{-1}$).
    Figure (b) displays the results for PI-DeepONet, showing a visually inaccurate prediction and a large absolute error (max 2.5).
    In contrast, Figure (c) shows the results for PO-CKAN, whose prediction is visually almost identical to the exact solution, with a maximum absolute error of only 0.016.
}
    \label{eikon_circle_result}
\end{figure}

\paragraph{\textbf{Results and Discussion}}
Figure~\ref{eikon_circle_result} demonstrates a clear performance gap, with PI-DeepONet unable to accurately approximate the operator, exhibiting elevated test loss and reduced prediction accuracy, likely due to its limited network depth. In contrast, PO-CKAN accurately reproduces the circular SDFs, achieving a mean relative $L^2$ error of $\mathbf{5.10 \times 10^{-3}}$, demonstrating its effectiveness in learning operators from geometric inputs.

\subsection{Fractional Partial Differential Equations}

To assess PO-CKAN’s capability in learning non-local, history-dependent operators, we next investigate a time-fractional partial differential equation (FPDE) characterized by variable-order derivatives~\cite{lu_fpinn-deeponet_2025}. Such equations are prevalent in modeling anomalous transport and memory effects in systems such as viscoelastic materials, porous media, and diffusion-wave processes.

The governing equation is defined on $t \in [0,1], x \in [0,\pi]$:
\begin{equation}
    \begin{cases}
        D_t^{\alpha(t)} u(x,t) = \frac{\partial^2 u}{\partial x^2} + f(x,t),\\
        f(x,t) = \frac{\Gamma(4)}{\Gamma(4-\alpha(t))} t^{3-\alpha(t)} \sin(x) + t^3 \sin(x),
    \end{cases}
\end{equation}
where $D_t^{\alpha(t)}$ denotes the variable-order Caputo fractional derivative, with a time-varying order $\alpha(t) \in [0,0.9]$, introducing additional temporal complexity~\cite{lu_fpinn-deeponet_2025}. 

Homogeneous initial and boundary conditions are imposed: ($u(x,0)=u(0,t)=u(\pi,t)=0$)

To facilitate validation, the method of manufactured solutions is used—constructing $f(x,t)$ such that the analytical solution is $u(x,t)=\sin(x) t^3$

\paragraph{\textbf{Network Architecture and Training Procedure}}
The PO-CKAN model consists of branch and trunk networks, each implemented as compact two-layer CKANs (rational degree \(n=4\), , chunk configuration $2 \times 2$), with 20 units per layer and a 40-dimensional latent output.

Training is conducted entirely via physics-informed supervision, without paired data, by minimizing the composite loss:
\begin{equation}
    \mathcal{L}(\theta) = \lambda_{ic}\mathcal{L}_{ic}(\theta) + \lambda_{r}\mathcal{L}_{r}(\theta), \, (\lambda_{ic}, \ \lambda_{r})=(1, 10),
\end{equation}
The PDE residual term penalizes deviations from the governing equation:
\begin{equation}
    R_\theta(x,t) = D_t^{\alpha(t)} G_\theta(u_0) - \frac{\partial^2 G_\theta(u_0)}{\partial x^2} - f(x,t).
\end{equation}
where the fractional derivative is numerically approximated at discrete time steps $t_k$ via the convolutional Caputo formula:
\begin{equation}
    D_t^{\alpha_k} u(x, t_k) \approx \frac{\tau^{-\alpha_k}}{\Gamma(2-\alpha_k)} \sum_{j=1}^{k} c_j^{\alpha_k} \big[u(x,t_{k-j+1}) - u(x,t_{k-j})\big],
\end{equation}
with $\tau$ as the temporal step size and $c_j^{\alpha_k}$ denoting convolution weights.

This formulation enables stable computation of the fractional residual while preserving differentiability for gradient-based training.

\begin{figure}[htbp]
    \centering
    \includegraphics[width=1\linewidth]{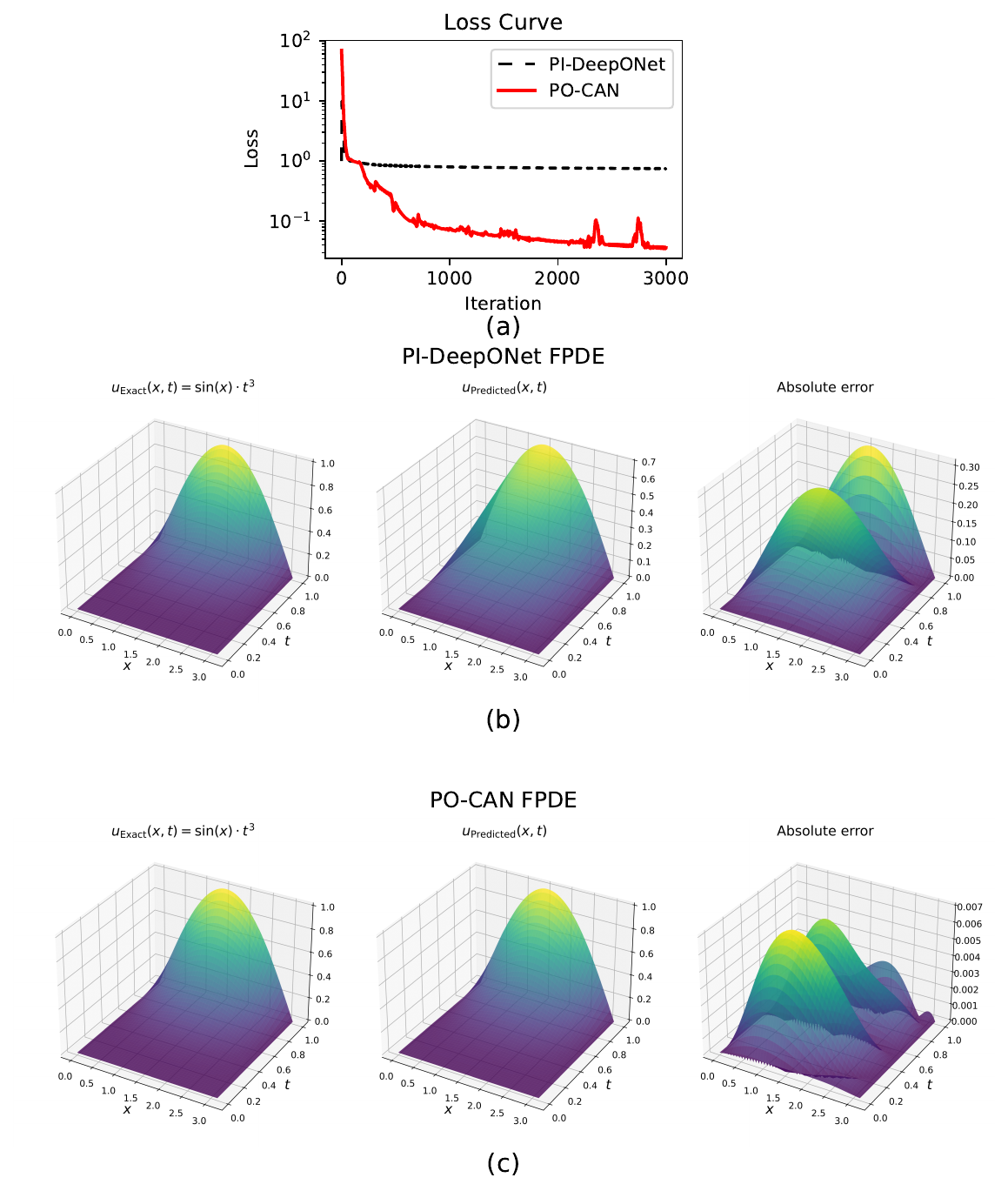}
    \caption{
    \textbf{Performance comparison of PO-CKAN and PI-DeepONet for solving a fractional PDE (FPDE).}
    The loss curve in (a) shows that while PI-DeepONet's loss remains high ($\sim 10^0$), PO-CKAN's loss has a low baseline ($\sim 10^{-1}$).
    The 3D plots show that PI-DeepONet's prediction remains inaccurate with a maximum error of 0.30 (b).
    In contrast, PO-CKAN's prediction of the solution $u(x, t) = \sin(x) \cdot t^3$ is visually correct, and its maximum absolute error is exceptionally low at 0.005.
}
    \label{fig:fractional_result}
\end{figure}

\paragraph{\textbf{Results and Interpretation}}
Figure~\ref{fig:fractional_result} highlights the model's strong performance on this challenging fractional problem. Quantitatively, PO-CKAN achieves a mean relative $L^2$ error of $\mathbf{2.54 \times 10^{-2}}$, representing an improvement of over 80\% compared to the baseline ($1.32 \times 10^{-1}$). Qualitative comparisons further confirm that our model closely reproduces the reference solution, whereas the baseline shows noticeable deviations. These results demonstrate the architecture's ability to accurately learn fractional operators and capture complex, history-dependent dynamics.

\subsection{Diffusion-reaction systems}
We finally assess PO-CKAN on a nonlinear diffusion–reaction equation, a class of problems that couples spatial diffusion with local chemical kinetics\cite{wang_learning_2021}.

Such systems are ubiquitous in biological pattern formation, chemical reactors, and ecological dynamics, serving as critical tests of both accuracy and stability for operator-learning frameworks. 

The governing equation reads:
\begin{equation}
\frac{\partial s}{\partial t} = D \frac{\partial^2 s}{\partial x^2} + k s^2 + u(x), \quad (x,t) \in (0,1] \times (0,1],
\label{eq:pde_diff_react}
\end{equation}
subject to homogeneous initial and boundary conditions, with diffusion coefficient $D=0.01$ and reaction rate $k=0.01$. 

\paragraph{\textbf{Network Architecture and Training Setup}}
The branch and trunk networks of PO-CKAN each comprise a four-layer CKAN (rational degree $n=4$,  chunk configuration $2 \times 2$) with 50 neurons per layer.

For comparison, PI-DeepONet uses identical layer widths but standard MLP blocks with tanh activations.

Both models are trained using the Adam optimizer for 120,000 terations with a composite loss 

\begin{equation}\label{DR-loss}
    \mathcal{L}(\theta) =  \lambda_{\text{ic}}\mathcal{L}_{\text{ic}}(\theta)+\lambda_{\text{bc}}\mathcal{L}_{\text{bc}}(\theta) + \lambda_{\text{r}}\mathcal{L}_{\text{r}}(\theta), \, (\lambda_{\textbf{ic}},\lambda_{\textbf{bc}}, \lambda_{\textbf{r}})=(1, 1, 1)
\end{equation}
where $\mathcal{L}_{\text{ic}}, \mathcal{L}_{\text{bc}}$ enforces the zero initial and boundary conditions, and $\mathcal{L}_{\text{r}}$ penalizes the PDE residual,
\begin{equation}
R^{(i)}_\theta(x,t) = \frac{\partial G_\theta(u^{(i)})}{\partial t} - D \frac{\partial^2 G_\theta(u^{(i)})}{\partial x^2} - k \big[G_\theta(u^{(i)})\big]^2 - u^{(i)}(x),
\label{eq:residual_diff_react}
\end{equation}
evaluated at collocation points across the spatio-temporal domain using automatic differentiation.

\begin{figure}[htbp]
    \centering
    \includegraphics[width=1\linewidth]{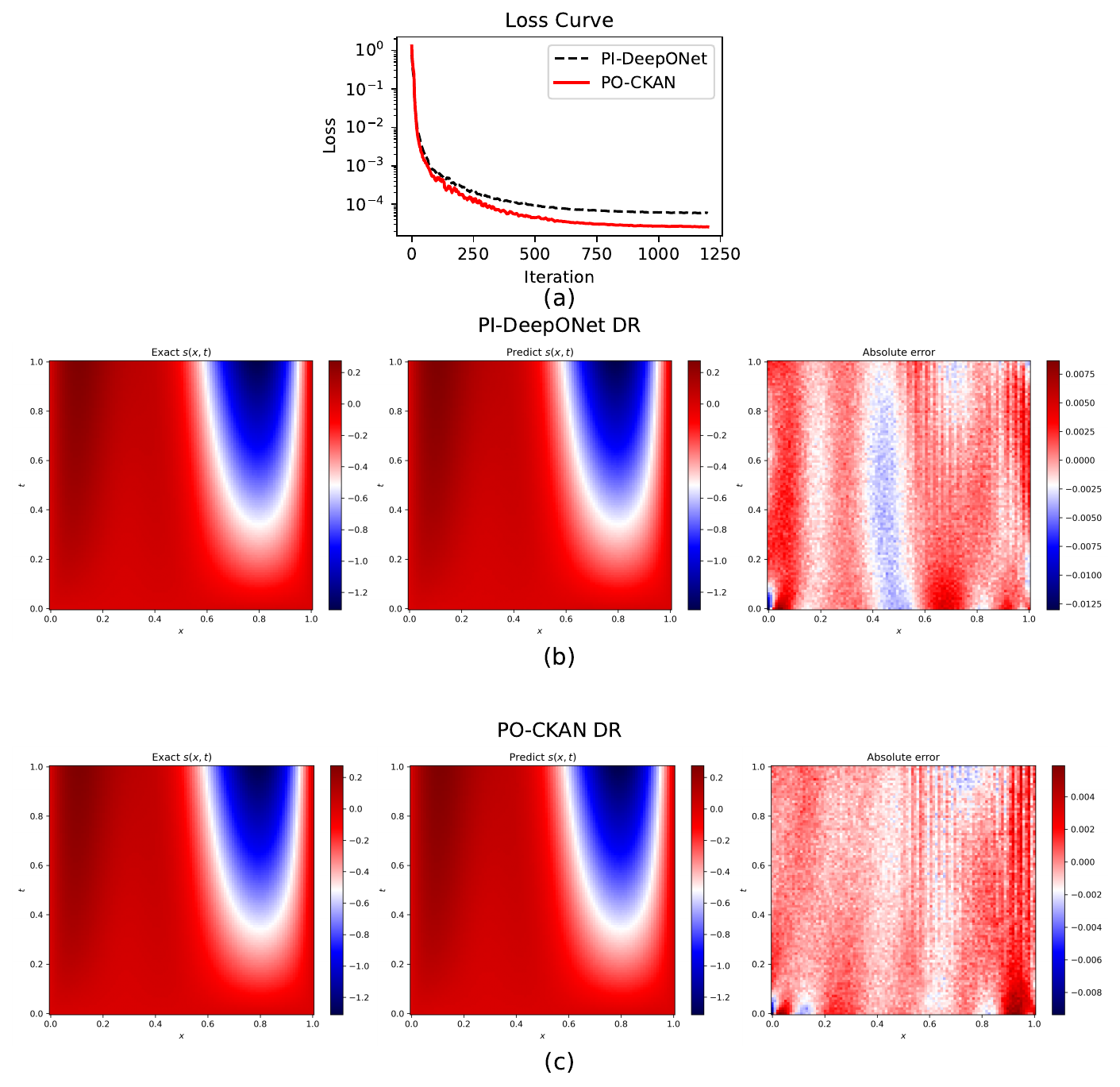}
\caption{
    \textbf{Performance comparison of PO-CKAN and PI-DeepONet on the diffusion-reaction (DR) problem.}
    The loss curves in (a) show that PO-CKAN reaches a final loss ($\sim 10^{-4}$) approximately an order of magnitude lower than PI-DeepONet ($\sim 10^{-3}$).
    The PI-DeepONet model provides a good visual prediction for the solution $s(x, t)$ (b), with a maximum absolute error of approximately 0.0275.
    The PO-CKAN model (c) further improves upon this result, yielding a nearly identical visual prediction and reducing the maximum absolute error by a factor of 7, to just 0.004.
}
    \label{fig:diffusion_reaction}
\end{figure}

Figure~\ref{fig:diffusion_reaction} illustrates the model performance. PO-CKAN exhibits faster convergence and a lower final loss compared to PI-DeepONet. Quantitatively, it achieves a mean relative $L^2$ error of $\mathbf{2.58 \times 10^{-3}}$, representing an improvement of over 50\% relative to the baseline ($5.19 \times 10^{-3}$). Solution plots further confirm that PO-CKAN predictions closely match the ground truth, while PI-DeepONet outputs display noticeable discrepancies. This benchmark underscores the effectiveness of the proposed architecture in learning operators for complex, nonlinear systems with high fidelity.

\subsection{Ablation Studies}\label{sec:ablation}

We conducted ablation studies to investigate the impact of two key architectural hyperparameters: (i) chunk granularity $c$ and (ii) ERU order $n$. For all experiments, we used identical data splits, optimizers, and training budgets to ensure a fair comparison.

\subsubsection{Effect of Chunk Granularity (c)}\label{sec:ablate_chunks}

Table \ref{tab:ablate_chunks} presents the model complexity and inference efficiency for different values of chunk granularity $c$. The results clearly demonstrate that as $c$ increases, both the model's parameter count and computational load (FLOPs) grow substantially. Specifically, increasing $c$ from 2 to 50 (equivalent to a full KAN) causes the number of parameters to expand by more than 8.5 times (from approximately 23k to 198k) and FLOPs to increase by 7.6 times. In contrast, the impact on inference time is far more moderate, with only a 39\% increase from 4.16 ms to 5.79 ms. This highlights a critical trade-off: while a finer chunk granularity (higher $c$) significantly raises the theoretical computational cost, its effect on practical inference latency is modest. This is expected, as higher granularity allows the model to capture more intricate relationships, potentially improving accuracy at the expense of increased model complexity.

\begin{table}[h]
\centering
\caption{Model complexity and inference efficiency for different values of $c$. The table reports the number of parameters, FLOPs, and average inference time per sample.}
\label{tab:ablate_chunks}
\begin{tabular}{lccc}
\toprule
$c$ & Params & FLOPs & Inference (ms) \\
\midrule
2 & 23230 & 62376 & 4.164 \\
5 & 24700 & 88140 & 4.392 \\
10 & 29950 & 131080 & 4.612 \\
25 & 66700 & 259900 & 5.140 \\
50 (\text{full KAN}) & 197950 & 474600 & 5.789 \\
\bottomrule
\end{tabular}
\end{table}

\subsubsection{Effect of ERU Order (n)}\label{sec:ablate_eru}

Table \ref{tab:ablate_eru} investigates the impact of the ERU order $n$ on model accuracy, with a fixed chunk granularity of $c=2$. The results show a clear trend: increasing $n$ leads to a consistent and significant reduction in the relative $L^2$ error. For instance, by raising $n$ from 4 to 20, the error decreases substantially from $6.23\times 10^{-2}$ to $1.19\times 10^{-2}$, an improvement of over 80\%. Notably, this substantial gain in accuracy comes at a negligible cost in model complexity. Over the same range of $n$, the parameter count increases minimally from 23,230 to 24,126, a rise of less than 4\%. This experiment powerfully demonstrates that increasing the ERU order is a highly parameter-efficient strategy for enhancing model performance, allowing for a major boost in function approximation capability with a minimal increase in model size.

\begin{table}[h]
\centering
\caption{Effect of ERU order $n$ with fixed $c=2$. The final column indicates the mean relative $L^2$ error.}
\label{tab:ablate_eru}
\begin{tabular}{lcc}
\toprule
$n$ & Params & Relative $L^2$ Error \\
\midrule
4 & 23230 & $6.23\times 10^{-2}$ \\
8 & 23454 & $2.33\times 10^{-2}$ \\
12 & 23678 & $2.26\times 10^{-2}$ \\
16 & 23902 & $1.27\times 10^{-2}$ \\
20 & 24126 & $1.19\times 10^{-2}$ \\
\bottomrule
\end{tabular}
\end{table}

\section{Conclusion and Future Work}
\label{sec:conclusion}
In this study, we proposed the Physics-Informed Deep Operator Chunk-rational Kolmogorov–Arnold Network (PO-CKAN)—a scalable and expressive neural-operator framework that combines physics-informed learning with chunk-wise rational Kolmogorov–Arnold representations. Through comprehensive experiments on Burgers’, Eikonal, fractional, and diffusion–reaction equations, we demonstrated that PO-CKAN consistently surpasses conventional PI-DeepONet baselines in accuracy, convergence speed, and numerical stability.

The key innovation lies in replacing conventional MLPs with the CKAN layer, which employs rational activation functions and parameter-shared chunk structures to overcome the quadratic parameter growth that limits standard KANs. This design enables efficient operator learning across nonlinear, nonlocal, and stiff PDEs while preserving interpretability and smooth functional mapping.

Our findings highlight several important outcomes:
\begin{itemize}
    \item Accuracy: PO-CKAN achieves up to 80 \% reduction in relative $L^2$ error across multiple PDE families.
    \item Efficiency: The chunk-wise sharing strategy and rational activations provide an order-of-magnitude reduction in computational cost.
\end{itemize}

Beyond performance gains, the PO-CKAN architecture provides a generalizable template for building scalable neural operators with embedded physical priors.

Its modular structure allows easy adaptation to emerging architectures such as transformer-based neural operators, Fourier or graph-based kernels, and hybrid symbolic–neural solvers.

Looking ahead, several research directions merit further exploration:

\begin{itemize}
    \item Extension to multi-physics coupling: applying PO-CKAN to coupled thermo-hydrodynamic or chemo-mechanical systems~\cite{ameya_d_jagtap_extended_2020}.
    \item Uncertainty quantification and Bayesian calibration: incorporating stochastic priors over CKAN activations to capture model uncertainty~\cite{yang_b-pinns_2021}.
    \item Hybrid data assimilation: integrating limited experimental observations with physics-based priors for data-scarce regimes.
    \item Hardware acceleration: exploiting sparsity and parallelism in chunked rational functions for efficient implementation on GPUs or photonic processors.
\end{itemize}

\section*{Acknowledgment}
This work was supported by the National Science Foundation (NSF) under grants DMS-2053746, DMS-2134209, ECCS-2328241, CBET-2347401 and OAC-2311848, and by the U.S.~Department of Energy (DOE) Office of Science Advanced Scientific Computing Research program under award number DE-SC0023161, and the DOE–Fusion Energy Science program, under grant number: DE-SC0024583.

\bibliographystyle{abbrv} 
\bibliography{references} 
\appendix

\end{document}